# ACCURATE AND EFFICIENT VIDEO DE-FENCING USING CONVOLUTIONAL NEURAL NETWORKS AND TEMPORAL INFORMATION

*Chen Du, Byeongkeun Kang, Zheng Xu, Ji Dai and Truong Nguyen*

Department of Electrical and Computer Engineering, UC San Diego, La Jolla, CA 92093 USA
{c9du, bkkang, zhx114, jid046, tqn001}@ucsd.edu

## ABSTRACT

De-fencing is to eliminate the captured fence on an image or a video, providing a clear view of the scene. It has been applied for many purposes including assisting photographers and improving the performance of computer vision algorithms such as object detection and recognition. However, the state-of-the-art de-fencing methods have limited performance caused by the difficulty of fence segmentation and also suffer from the motion of the camera or objects. To overcome these problems, we propose a novel method consisting of segmentation using convolutional neural networks and a fast/robust recovery algorithm. The segmentation algorithm using convolutional neural network achieves significant improvement in the accuracy of fence segmentation. The recovery algorithm using optical flow produces plausible de-fenced images and videos. The proposed method is experimented on both our diverse and complex dataset and publicly available datasets. The experimental results demonstrate that the proposed method achieves the state-of-the-art performance for both segmentation and content recovery.

## 1. INTRODUCTION

With the popularity of smartphones and affordable cameras, photography has become a universal method to record and share memorable moments. However, in many places such as a zoo, a playground, or a construction site, the views of cameras are obstructed by fences that secure people from potential risks. The obstructed view by a fence often destroys the aesthetic experience of photographers and limits the accurate recognition of objects behind the fence. Hence, an efficient de-fencing method is demanded not only to assist photographers but also to improve the ability of computer vision algorithms such as object detection and recognition.

The general pipeline for the state-of-the-art de-fencing methods includes two main steps: fence segmentation and content recovery. In fence segmentation, the fence is detected, segmented, and removed from an image. In content recovery, the eliminated area is recovered with plausible content.

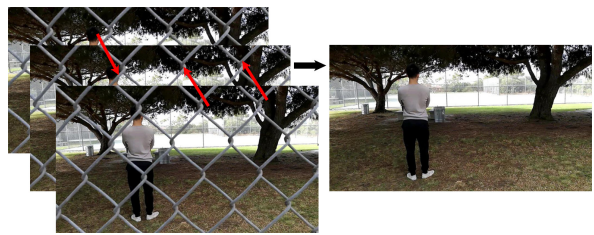

**Fig. 1**. Video de-fencing. Left image and right image show the input and the output of the de-fencing system. Red arrows on the left image show the recovery of content from neighboring frames to a target frame.

### 1.1. Related Work

**Fence segmentation.** Several fence segmentation methods have been proposed in recent years. Park *et al.* proposed a near-regular texture detection method [1] for fence segmentation. This method performs well for regular fence with uniform background. However, near-regular texture detection is still challenging in real-world scenes with various lighting conditions and complex background. Khasare *et al.* applied image matting [2] to segment a fence from background [3]. This method requires scribbles along the fence from a user, limiting its efficiency. To avoid scribbles from humans, Jonna *et al.* proposed a deep learning-based method for fence segmentation [4–6]. The main idea is detecting the joints of fences using convolutional neural networks (CNN) and support vector machine (SVM) and then connecting the joints to obtain scribbles for image matting. Although this method has relatively better efficiency and robustness, it has limitations in segmenting fences with irregular patterns.

For the segmentation of arbitrarily shaped fences, a motion-based approach was proposed in [7] with the hypothesis of a static background. Yi *et al.* proposed a fence segmentation method using optical flow and graph-cut with a spatiotemporal refinement [8]. This method performs well for a moving camera and irregular fences using both motion and color information. However, this method cannot detect fences using a single frame and has high computational complex-ity. In [9], Jonna *et al.* segmented fence based on disparity estimation. The disparity was estimated using a pre-trained CNN [10] that processes stereo matching on image pairs. Although this method can handle irregular fences, it requires stereo image pairs and hence, is inconvenient.

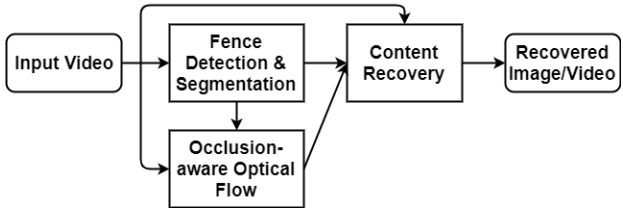

**Fig. 2**. Overview of the proposed de-fencing system.

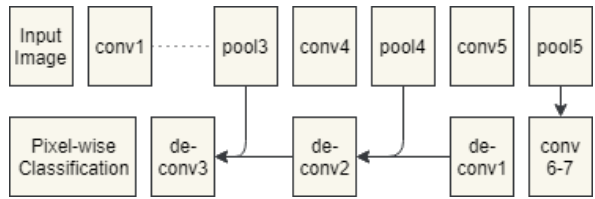

**Fig. 3**. Architecture of the FCN-8s [13].

**Content recovery.** Given fence segmentation, the next step is recovering the content of the eliminated areas. Conventionally, content recovery methods employed image inpainting methods [8, 11]. However, these methods have limitations in recovering high-frequency details and in applying for video de-fencing because of the potential discontinuity around edges between adjacent frames.

For video de-fencing, the occluded content in a frame can be recovered from neighboring frames as shown in Fig. 1. It is processed by estimating the relative motion between a target frame and adjacent frames and by recovering content using the motion and the backgrounds in contiguous frames. To simplify the motion estimation, Khasare *et al.* [3] and Mu *et al.* [12] assumed that the background pixels in the frames shift equally. However, it is barely valid in complex scenes with objects at various distances. In [6], Jonna *et al.* proposed a robust method for background motion estimation by introducing occlusion-aware optical flow. Along with the robust motion estimation, they fused multiple frames with L1-norm regularization. This method efficiently preserves the background details. However, it requires users to manually select the frames and suffers from over-smoothing.

In this paper, we propose a fence segmentation method using the fully convolutional neural networks and temporal refinement. The segmentation method achieves significant improvements in both accuracy and efficiency. It also does not require any manual labeling and is able to handle irregular fences and challenging scenes such as having similar color for both background and fences. Moreover, we propose a fast and robust recovery method using occlusion-aware optical flow [6]. The recovery method produces more plausible de-fenced images and videos. Lastly, we demonstrate the proposed method achieves the state-of-the-art performance on both publicly available dataset and our novel dataset which contains diverse and complex scenes with accurate ground truths for fence segmentation. Fig. 2 shows the structure of the proposed de-fencing system.

This paper is organized as follows. We outline the details of the proposed fence segmentation method in Section 2. In Section 3, we introduce the content recovery method. The experiment results are presented in Section 4. Conclusion is described in Section 5.

## 2. FENCE SEGMENTATION

Fence segmentation is required in de-fencing to eliminate the pixels of fences so that the pixels can be recovered with the occluded content. Assuming content recovery algorithm can recover the eliminated pixels with plausible content, minimizing false negative errors is more important than minimizing false positive errors. Especially for video de-fencing, since backgrounds are mostly consistent in a short period, the hypothesis is valid in general. Still, minimizing false positive errors is also important since content recovery algorithm is not ideally precise because of the noises from motion estimation, light condition variation, etc.

### 2.1. Fully Convolutional Neural Network

To achieve accurate fence segmentation for high-quality de-fencing results, we approach this task using the fully convolutional neural network (FCN) [13]. We employ this network since it has demonstrated the state-of-the-art performance in semantic segmentation task. The network replaces fully connected layers in VGG-16 network [14] by convolutional layers to preserve spatial information. Also, the network attaches deconvolution layers to compensate pooling layers and to obtain the output with the same resolution of the input. Moreover, the network adds skip connections to achieve refinement and denser predictions. The architecture is shown in Fig. 3.

In training, we randomly sample images and augment them by cropping, scaling, rotating, flipping, adding noise, and distorting HSV value to improve the robustness of the trained model. We initialize FCN-32s network using the VGG-16 model which is pre-trained on PASCAL VOC dataset [15]. Then, we sequentially train FCN-32s, FCN-16s, and FCN-8s using the fixed learning rate of $10^{-10}$, the momentum of 0.99, and the weight decay of $5 \times 10^{-4}$. The maximum number of iterations is $1.5 \times 10^5$ for FCN-32s and FCN-16s, and $9 \times 10^5$ for FCN-8s. The trained FCN-8s is used to segment fences. Sample predictions using the trained network are shown in Fig. 4 (b).

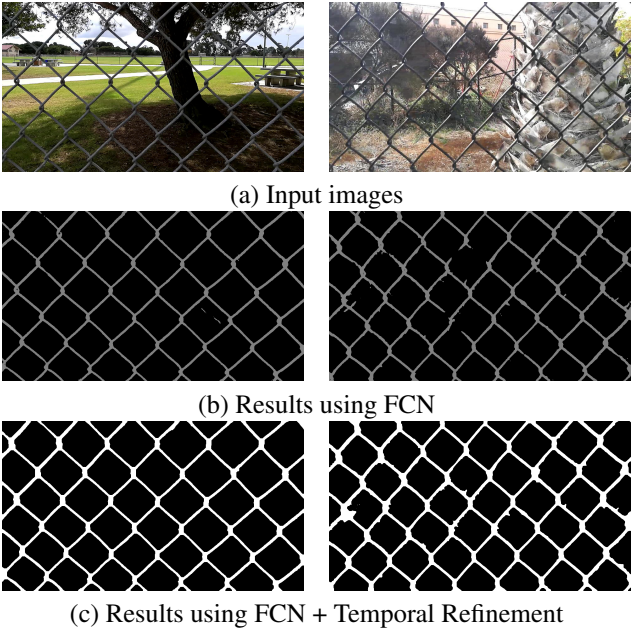

(a) Input images

(b) Results using FCN

(c) Results using FCN + Temporal Refinement

**Fig. 4**. Results of the proposed fence segmentation method.

## 2.2. Temporal Refinement

We propose a temporal refinement method to correct misclassified pixels using temporal information. The purpose is to remove fences completely at every frame. To get the refined fence prediction $P'_o$ on a target frame $o$, we first estimate the geometric transformation $T_{ko}$ from the neighboring frame $k$ to the target frame using phase correlation method [16]. We, then, warp the neighboring frame to the target frame. This is repeated for $m$ adjacent frames, and we average the warped predictions $T_{ko}P_k$. To prevent the accumulation of false positive errors, only the pixels whose averaged prediction score is higher than the preset threshold ($\mu$) are determined as valid fence pixels. Finally, we process logical OR operation ($||$) between the prediction $P_o$ on the target frame and the thresholded prediction of neighboring frames (see (1)).

$$P'_o = P_o || \Big(\frac{1}{m}\sum_{k=1}^{m} T_{ko}P_k \geq \mu\Big). \quad (1)$$

Minor refinements are further processed using morphological closing operations. Fig. 4 (c) shows that the temporal refinement can efficiently reduce false negative errors.

## 3. CONTENT RECOVERY

### 3.1. Background Motion Estimation

To recover the occluded background by a fence, we use the occlusion-aware optical flow [6] which is improved from the coarse-to-fine optical flow framework [17]. Let $Y_o$ and $Y_k$ refer to the target image and the $k^{th}$ neighboring image, respectively. The increment of the optical flow is computed as follows:

$$E(du, dv) = \operatorname*{argmin}_{dF} \|\neg P'(M_{F+dF}Y_k - Y_o)\|_1 \\ + \lambda(\|\nabla(u+du)\|_1 + \|\nabla(v+dv)\|_1) \quad (2)$$

where $F = [u, v]$ is the currently estimated optical flow; $M_{F+dF}$ is the warp matrix with increment $dF$; $\nabla$ is the gradient operator; $\lambda$ is the weight between data cost term and regularization loss term; $P' = P'_o || M_{F+dF}P'_k$ is the output of logical OR operation between fence predictions in the target frame $o$ and the warped $k^{th}$ frame; $P$ is a boolean matrix that true value represents fence, and negating it switches between true and false. By ignoring the data cost term, the increment of the optical flow of occluded pixels is computed only based on the regularization term. Hence, it provides the motion estimation of the fenced pixels. We apply successive-over-relaxation (SOR) algorithm using iterative re-weighted least squares (IRLS) method to solve for $dF$.

### 3.2. Data Fusion

Given the estimated optical flow of background, we need to recover plausible contents from adjacent frames. This process consists of selecting meaningful frames and determining the content at each pixel. For selecting frames, [6] manually selected the frames for simplicity. However, it is not feasible for video de-fencing. Hence, we use $n$ contiguous frames with weights. When deciding $n$, incorporating more frames has the advantage of obtaining abundant background information. However, it could also increase noises because of acquiring information from distant frames. To suppress the noises, we estimate contents using the weighted mean $\hat{X}_o$ of selected $n$ frames. For determining the content, we recover the content with the RGB value in a specific frame where the value is the closest value among $n$ adjacent frames to the weighted mean $\hat{X}_o$. This is to avoid over-smoothing and to preserve high-frequency details.

To compute the weighted mean $\hat{X}_o$, we use the total variance (TV)-L1 de-noising model [18] as follows:

$$\hat{X}_o = \operatorname*{argmin}_{X} \left\| X - \sum_{k=1}^{n} w_k M_{ko} \neg P'_k Y_k \right\|_2^2 + \lambda \|\nabla X\|_1 \quad (3)$$

where $w_k$ and $M_{ko}$ represent the weight and the warp matrix for the adjacent frame $Y_k$, respectively. The weight parameter $w_k$ is negatively correlated with the temporal distance between $Y_k$ and the target frame $Y_o$. Also, $w_k$ is normalized to ensure $\sum_{k=1}^{n} w_k = 1$. The warp matrix $M_{ko}$ from $Y_k$ to $Y_o$ is obtained by the occlusion-aware optical flow. By multiplying the negative of fence prediction $P'_k$, only the non-fence pixels are taken into account in the L2-loss term.

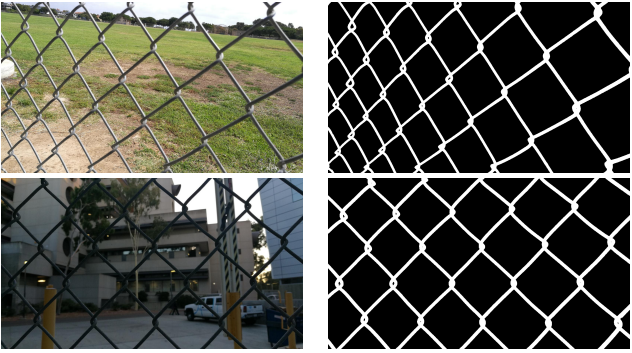

**Fig. 5**. Fence segmentation dataset.

Since directly recovering using the weighted mean $\hat{\boldsymbol{X}}_o$ causes an over-smoothing problem such as blurred boundaries because of averaging over multiple frames, we recover the content with the nearest RGB value in a specific frame among all candidate frames. For each pixel $(x, y)$, We first search the index $\boldsymbol{I}$ of the warped frame that has the non-fence pixel at $(x, y)$ with nearest RGB value as follows:

$$\boldsymbol{I}_o(x,y) = \underset{i}{\operatorname{argmin}} \left\| \hat{\boldsymbol{X}}_o(x,y) - \boldsymbol{M}_{io}\boldsymbol{Y}_i(x,y) \right\|_2^2 \quad (4)$$
$$s.t. \quad \boldsymbol{M}_{io}\boldsymbol{P}'_i(x,y) = \text{false}$$

Then, the real RGB value are used to get the plausible content $\boldsymbol{X}'_o$.

$$\boldsymbol{X}'_o(x,y) = \boldsymbol{M}_{\boldsymbol{I}_o(x,y)o}\boldsymbol{Y}_{\boldsymbol{I}_o(x,y)}(x,y) \quad (5)$$

Finally, we substitute the fence pixels in the target frame with the mapped RGB values to get the de-fenced result $\boldsymbol{X}_o$.

$$\boldsymbol{X}_o = \neg \boldsymbol{P}'_o \boldsymbol{Y}_o + \boldsymbol{P}'_o \boldsymbol{X}'_o \quad (6)$$

The solution of the TV-L1 problem can be efficiently approximated using the proximal TV method [19]. For those rare pixels that are occluded in all selected frames, the inpainting [20] is used to predict the content.

## 4. EXPERIMENT RESULTS

We first introduce our novel fence segmentation dataset containing diverse and complex scenes in Section 4.1. Then, we evaluate the proposed fence segmentation method on both publicly available datasets and our dataset, comparing to the other state-of-art methods [1, 9] in Section 4.2. Lastly, we demonstrate the performance of the proposed de-fencing method on videos in Section 4.3. We also compare the results with [6] on different datasets. For hyperparameters, we choose $m = 5$ for temporal refinement, $n = 4$–$8$ for content recovery, and $\lambda = 0.0005$ for regularization. All the experiments are performed using a machine with Intel Core i7 4.20GHz processor and Nvidia GTX 1080 graphics card. The proposed de-fencing method with $n = 4$ takes less than 10 seconds to process a frame with the resolution of $960 \times 540$.

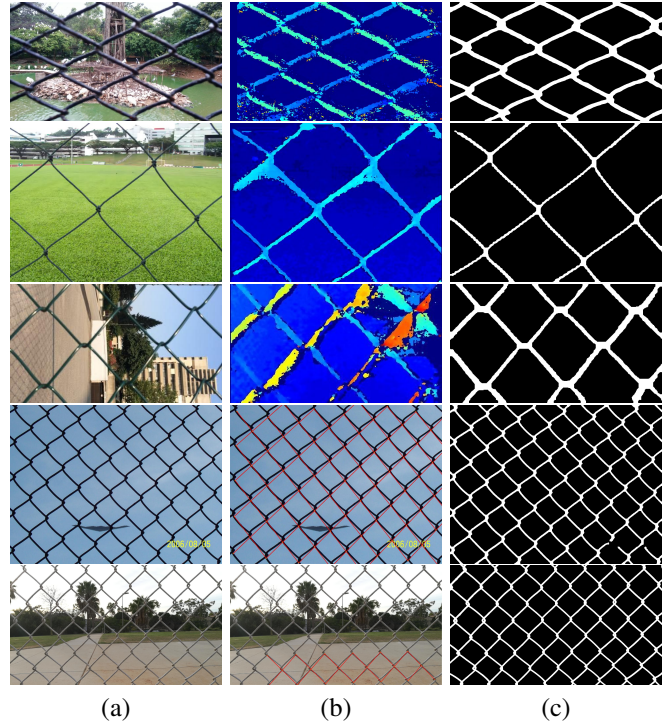

(a)          (b)          (c)

**Fig. 6**. Comparison of fence segmentation. (a) input images; (b) estimated disparity by [9] for the row 1–3 and detection results by [1] for the row 4–5; (c) results by the proposed method. Input images are from [1, 8, 9, 12] and our test set.

### 4.1. Fence Segmentation Dataset

We collected a novel dataset containing diverse and complex scenes with fences. Although several large datasets exist for semantic segmentation [15, 21, 22], most of the fences in these datasets are treated as a single blob or are labeled roughly. Thus, the datasets cannot be used in de-fencing which needs precise ground truth labels in pixel-level. Therefore, we collected our own dataset for real-world fence segmentation[1].

The dataset consists of 645 fence images captured at various locations and in diverse light conditions. The resolution of each image is 3264×1840. To obtain highly precise ground truth labels, we stabilize the camera and capture two images. One image is captured as usual. The other image is captured while we put a green curtain behind the fence. Then, we use color-based segmentation on the image with the green curtain to obtain the ground truth label. Several samples of the dataset are shown in Fig. 5.

### 4.2. Fence Segmentation Evaluation

We compare the proposed fence segmentation algorithm to the other state-of-the-art methods [1, 9] on both our test dataset and the datasets from [1, 8, 9, 12]. The qualitative results and the quantitative results are shown in Fig. 6 and Table 1. Since [9] requires stereo image pairs to estimate disparity and then to segment fences, we show their results on the datasets [8, 9, 12] that have image pairs. On the row 1–3 in Fig. 6, (a), (b), and (c) show one of the stereo pairs, the results of [9], and those of our method. As observed in all three images, the proposed method achieves more accurate results. The comparison with [1] is shown on the row 4–5. The results demonstrate that the proposed method is more robust to irregular pattern and complex background while [1] suffers from them. The quantitative evaluation on our test dataset is shown in Table 1, which confirms that our method has outstanding accuracy on real-world fence segmentation.

---
[1]The dataset is available at https://github.com/chen-du/De-fencing

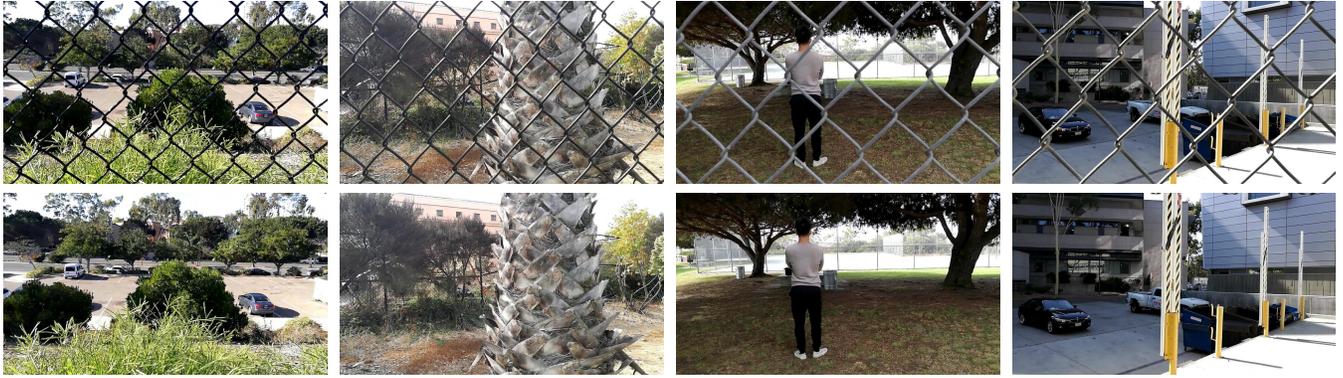

**Fig. 7**. De-fencing results on phone captured videos.

**Table 1**. Quantitative evaluation of fence segmentation

| Method | Precision | Recall | F-measure |
|---|---|---|---|
| Park *et al.* [1] | 0.500 | 0.163 | 0.246 |
| Proposed | 0.910 | 0.959 | 0.934 |

### 4.3. De-fencing evaluation

Visual de-fenced results on phone captured videos are shown in Fig. 7. The results show that our de-fencing method is able to remove the fence pixels accurately and recover the content clearly with high-frequency details. We compare our content recovery method to the FISTA optimization method [6] using the same fence segmentation in Fig. 8 (a), and compare the whole de-fencing system to [6] in Fig. 8 (b). The first row and the second row in Fig. 8 show their results and our results, respectively. In the third row, we enlarge a small region of their results and our results on the left and on the right, respectively. The overall results demonstrate that the proposed method outperforms other methods by recovering results with high-frequency details and sharper boundaries. The improvements are due to the weighted mean on neighboring frames and the use of real RGB values.

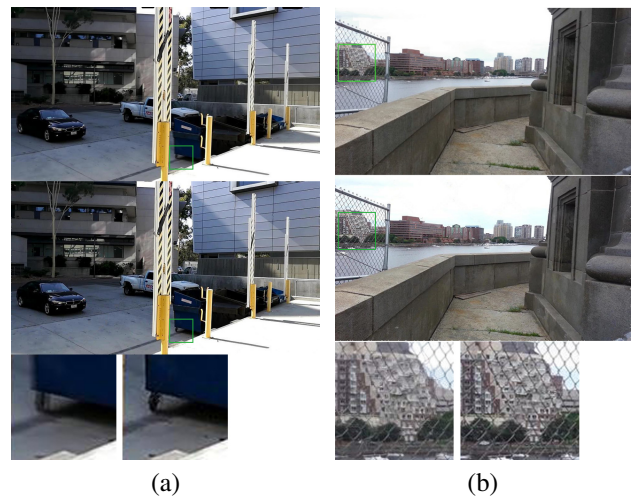

(a)      (b)

**Fig. 8**. Comparison of de-fencing results. Images in the first row and in the second row are the results by [6] and by the proposed algorithm, respectively. The left and right images in the third row show the zoom-in patches of the first row and the second row, respectively.

### 5. CONCLUSION

In this paper, we propose a novel de-fencing system consisting of deep learning-based segmentation and a fast/robust recovery algorithm for real-world images and videos. In segmentation, we propose an automatic fence segmentation algorithm using the fully convolutional neural network and temporal refinement. For content recovery, we estimate motion using the occlusion-aware optical flow and formulate data fusion using the framework which can be solved using the proximal TV. We evaluate the proposed method on both publicly available datasets and our diverse and complex fence segmentation dataset. The results demonstrate that the proposed method outperforms the other state-of-the-art methods in both fence segmentation and content recovery.